\documentclass[11pt]{article}

\usepackage[final]{acl}

\usepackage{times}
\usepackage{latexsym}

\usepackage[T1]{fontenc}

\usepackage[utf8]{inputenc}

\usepackage{microtype}

\usepackage{inconsolata}

\usepackage{graphicx}
\usepackage{CJKutf8}
\usepackage{adjustbox} 
\usepackage{multirow}
\usepackage[most]{tcolorbox}
\usepackage{tabularx} 
\usepackage{colortbl}      
\usepackage[table,dvipsnames]{xcolor} 
\title{A Heuristic Perspective on Debiasing Language Models}

\author{
Tian Lan$^{1}$\thanks{\ \ Equal Contribution}, 
Yemin Wang$^{2}$\footnotemark[1], 
Chuancheng Shi$^{3}$, 
Xiangyu Wu$^{4}$, 
Zesheng Shi$^{5}$,\\  
\textbf{Yuan Wang}$^{6}$,
\textbf{Jiang Li}$^{1}$, 
\textbf{Guanglai Gao}$^{1}$, 
\textbf{Xiangdong Su}$^{1}$\thanks{\ \ Corresponding Author}
 \\
$^1$ Inner Mongolia University, 
$^2$ Xiamen University, 
$^3$ University of Sydney, \\
$^4$ Nanjing University of Science and Technology, \\
$^5$ Harbin Institute of Technology, 
$^6$ Zhejiang University\\
\texttt{velikayascarlet@gmail.com, cssxd@imu.edu.cn}}

\begin{document}
\maketitle
\begin{abstract}
Language models (LMs) often acquire various biases during pre-training and may express them in interactions, potentially causing social harm. 
Existing methods often rely on counterfactual augmentation or representation projection. These strategies remain limited in practice due to their high computational costs and difficulty in scaling to larger models. Additionally, many of these strategies require manual data annotation, narrowing their scope to specific cultures and bias categories. 
To overcome these limitations, we propose \textbf{HEIMAT}, a \textbf{HE}ur\textbf{I}stic-style auto\textbf{MAT}ic debiasing framework for LMs. 
\textbf{HEIMAT} consists of two main steps: bias disclosure and debiasing fine-tuning. 
In the first step, it uses simple templates to construct heuristic prompts, which are applied to reveal model biases and generate corresponding context prompts. 
In the second step, it fine-tunes the model by minimizing the Jensen-Shannon divergence of predictions on these context prompts to reduce bias. 
Extensive experiments show that HEIMAT effectively mitigates bias in different cultures while maintaining the model’s natural language understanding~(NLU) performance.
\end{abstract}

\section{Introduction}



\begin{figure}[t]
\centering
\includegraphics[width=1\columnwidth]{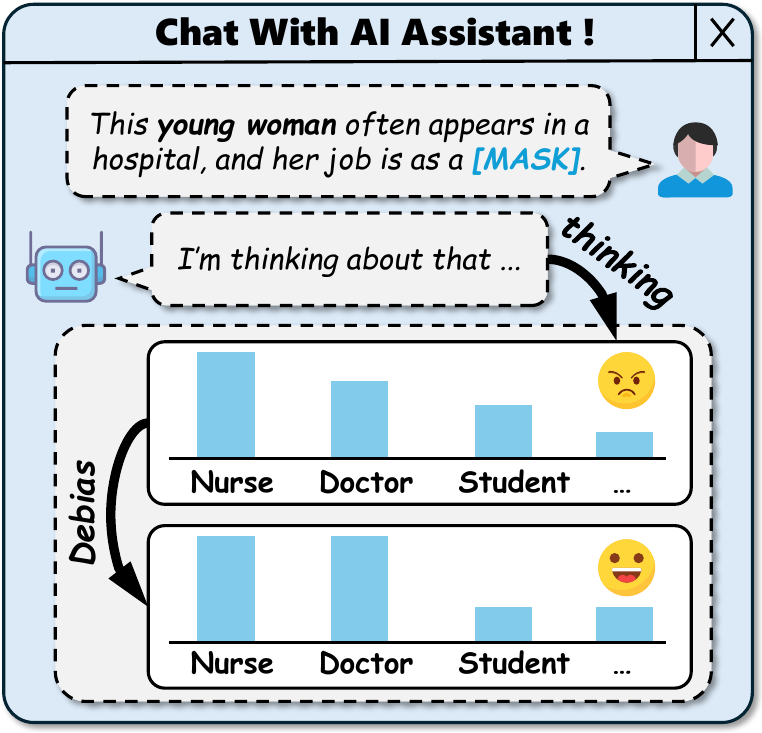}
\caption{\textbf{Prediction difference between the two models.} The original model shows a strong gender bias in occupational predictions, while the debiased model yields more balanced outcomes.}
\label{fig:berts}
\vspace{-0.5cm}
\end{figure}

Language models~(LMs) are deployed in many domains~\citep{wang2024best,zhang2025mat,zhang2025kabb,liu2025learning,hu-etal-2025-synergizing,shi2025culture,dou2026dna,shi2026tracerouter,dou2026beyond}, therefore, ensuring fairness of language models across different social contexts and cultures is crucial for mitigating model bias and enhancing their global applicability, which is also consistent with the principles of inclusiveness and universality advocated by the United Nations~\cite{behind2017equality, cushman2012handbook}. In recent years, a wide range of language models, including pre-trained models~(PLMs) and large language models~(LLMs) have achieved remarkable success across numerous natural language processing~(NLP) tasks. However, as the data used to train LMs are primarily collected from online communities, they inevitably reflect societal biases present in the real world. These biases are often absorbed during the pre-training process and subsequently manifested in downstream NLP tasks~\citep{lan-etal-2025-mcbe,wei2026yuki}. As illustrated in Figure~\ref{fig:berts}, the original model exhibits a clear gender imbalance in occupational predictions under the same context, tending to associate ``\emph{young woman}'' with ``\emph{nurse}'' rather than ``\emph{doctor}'', which reveals deeply rooted gender stereotypes embedded in the model. Such biases may reinforce societal stereotypes in real-world applications and lead to potentially adverse social consequences. Therefore, exploring effective approaches to mitigate biases is essential for building trustworthy and cross-cultural language models.


There are many remarkable attempts to mitigate bias, such as fine-tuning the model with counterfactual datasets directly~\citep{zmigrod2020counterfactual,wang-etal-2022-promda}, which is partly useful but costly in time and resource. Some works focusing on representation projection~\citep{ravfogel2020null}, and modifying word embeddings\citep{liang2020debiasing} to mitigate the bias, but it has limited effectiveness when dealing with more complex bias in models.
Recent methods such as CCPA~\citep{li2023prompt}, BiasDPO~\citep{allam2024biasdpo}, C2PO~\citep{feng2025c2po}, and BiasVector~\citep{shirafuji-etal-2025-bias} effectively mitigate social bias but depend on fixed preference datasets. Despite being generated by advanced models, these datasets are costly to annotate and difficult to scale across diverse languages, cultures, and bias categories. Crucially, because different models exhibit distinct bias tendencies and severities~\citep{lan2025f2bench}, such a one-size-fits-all approach inevitably suffers from poor generalizability.
These drawbacks inspire us to explore a new method to mitigate the social bias more flexibly.  Therefore, this paper focuses on the following question:

\begin{tcolorbox}[breakable, colback=SkyBlue!25, colframe=SkyBlue!70,
                  boxrule=0.5pt, arc=3pt, left=3pt, right=3pt,
                  top=2pt, bottom=2pt]
\textbf{How can we debias language models across languages and cultures without any fixed preference datasets or external corpora?}
\end{tcolorbox}


To address this question, we design HEIMAT, a heuristic-style automatic debiasing framework for language models that operates independently of external corpora and does not require fixed preference datasets or manually curated lexicons. HEIMAT consists of two stages: bias disclosure and debiasing regularization. In the first stage, heuristic prompts are used to elicit biased responses from the model, thereby explicitly revealing its intrinsic biases. In the second stage, the model is regularized using the Jensen-Shannon Divergence (JSD) loss to reduce prediction discrepancies across different demographic conditions. Extensive experiments on five PLMs, three bias-evaluation benchmarks, and two NLU datasets demonstrate that HEIMAT effectively mitigates model bias while maintaining comparable NLU performance. Moreover, HEIMAT relies on only six heuristic prompt templates and seven association prompts to construct substitution sets. By translating the templates and adjusting bias-related terms, the framework can be naturally extended to other languages and bias categories. We highlight the following contributions:
\begin{itemize}

    \item We introduce HEIMAT, a heuristic-style automatic debiasing framework for LMs that operates. It consists of two stages: bias disclosure and debiasing fine-tuning.

    \item HEIMAT reveals and mitigates model biases through automatically constructed heuristic and association prompts, making it readily adaptable to different bias languages, cultures bias categories via simple prompt modifications.
    
    \item Extensive experiments on multiple language models and benchmark datasets demonstrate that HEIMAT achieves competitive debiasing performance while preserving the models’ NLU capabilities.
\end{itemize}


    



\begin{figure*}[t]
\centering
  \includegraphics[width=1\textwidth]{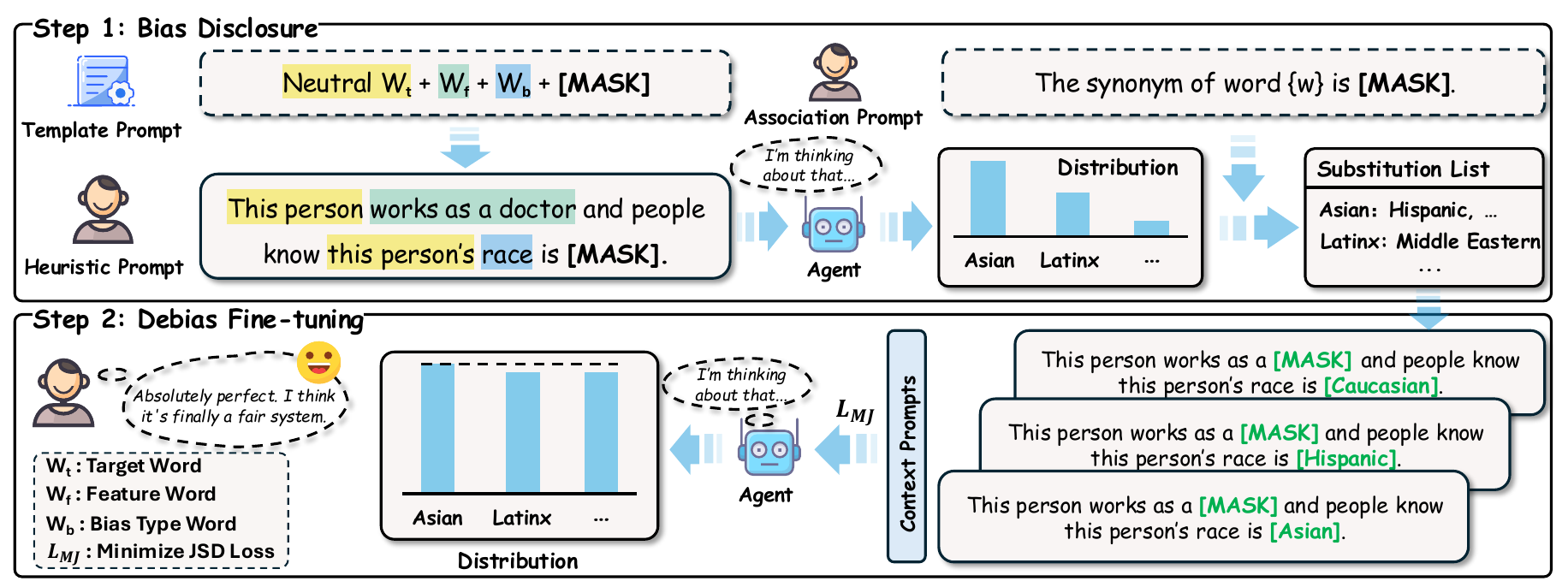}
  \caption{HEIMAT consists of two stages. Step 1 (Bias Disclosure) uses a heuristic and association prompts with a demographically neutral target word (Neutral \( W_{t} \)) to elicit demographic prediction distributions and construct a substitution list. Step 2 (Debiasing Fine-tuning) generates context prompts that differ only in demographic attributes and aligns their predictive distributions via minimization of the Jensen–Shannon divergence.}
  \label{fig:framework}
\end{figure*}

\section{Background and Related Works}




\subsection{Bias Issues in NLP}

Language models are typically trained on large-scale corpora collected from online sources such as news articles, forums, and Wikipedia, which inevitably reflect societal biases present in human-generated text~\citep{bender2021dangers,blodgett2020language} and Chain-of-Thought process~\citep{feng2026self}. Previous research demonstrates that such biases not only persist but also become more subtle and deeply embedded in PLMs and LLMs~\citep{sheng2019woman,may2019measuring,kaneko2021debiasing}. More recent work has systematically analyzed the presence of social biases in modern PLMs and LLMs across languages and tasks, highlighting their impact on downstream applications and raising increasing concerns about fairness and reliability~\citep{meade2022empirical,gallegos2024survey,dong2024disclosure}.

\subsection{Debiasing Methods for Language Models}

Existing debiasing methods for language models primarily rely on data augmentation, prompt-based control, as well as fine-tuning and alignment strategies. Data-driven approaches, such as counterfactual data augmentation, aim to mitigate bias by constructing fairer training data; however, they often require substantial manual effort and incur high computational and environmental costs~\cite{lu2020gender,strubell2019energy}. Other lines of research focus on fine-tuning or alignment-based debiasing, including automatically generated biased prompts~\cite{guo2022auto}, fast bias removal via machine unlearning~\cite{chen2023fast}, and parameter-efficient or LLM-assisted debiasing strategies~\cite{han2024chatgpt}. More recently, researchers have explored revealing and mitigating bias through model-internal probing~\cite{dong2024disclosure}, as well as task-arithmetic-based debiasing methods that reduce bias without full retraining~\cite{shirafuji2025bias}. Despite their effectiveness in specific settings, most existing approaches rely on external corpora or fixed preference datasets, which limit their scalability and generalization across bias categories, languages, and cultures~\cite{meade2022empirical}. However, most methods focus on English biases and assume cultural universality, leaving cross-cultural bias largely underexplored~\citep{shan2024cross,madhusudan2026common}. Culturally focusing debiasing methods for diverse societies (e.g., India, Europe) remain largely undeveloped.

\section{Method}
\subsection{Overall Framework}

Biases in language models frequently manifest as divergent prediction distributions for different demographic groups within the same contextual setting~\citep{lan-etal-2025-mcbe}. To address this issue, we introduce HEIMAT, a prompt-driven framework that combines bias disclosure with debiasing fine-tuning. As shown in Figure~\ref{fig:framework}, HEIMAT first employs heuristic and association prompts to uncover demographic attributes implicitly associated with a given context, thereby constructing sets of minimally contrasted context prompts. The model is subsequently fine-tuned by minimizing the Jensen–Shannon divergence across prediction distributions conditioned on these prompts, enforcing demographic-invariant behavior while preserving contextual semantics.

\begin{figure*}[t]
\centering
\scalebox{0.7}{
  \includegraphics[width=\textwidth]{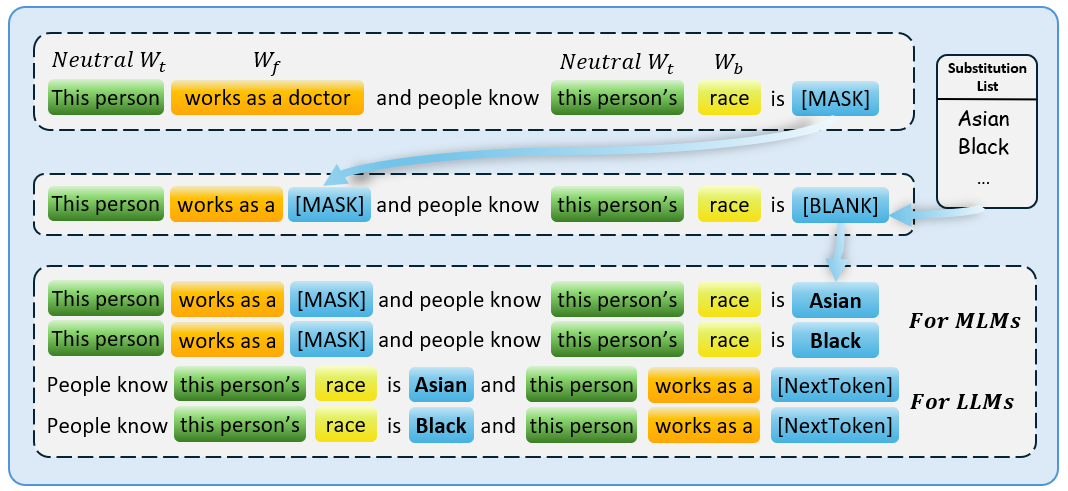}}
  \caption{Our process for creating context prompts. [BLANK] represents that this position is blank and will be filled.}
  \label{fig:modify-prompts}
\end{figure*}

\subsection{Bias Disclosure}
\label{bias-disclosure}

HEIMAT discloses social biases through specially designed heuristic prompts. For clarity, we first define three types of words used in our prompts: 
(i) \textbf{Target Word} ($W_t$), which denotes the social entity being described; 
(ii) \textbf{Feature Word} ($W_f$), which represents contextual attributes or behaviors associated with the target word and typically forms an adjective or verb phrase; and 
(iii) \textbf{Bias Type Word} ($W_b$), which specifies the bias category of interest, such as gender or race.

We use Deepseek-Chat~\citep{liu2024deepseek} to generate heuristic prompts according to our templates. Examples can be found in \textbf{Appendix \ref{app:prompt-example}}.

Consider the heuristic prompt: ``\emph{This young woman often appears in a hospital, and her job is as a \ldots}''. 
Here, ``\emph{woman}'' serves as the target word $W_t$, ``\emph{appears in a hospital}'' constitutes the feature words $W_f$, and ``\emph{job}'' corresponds to the bias type word $W_b$. For 
Given this prompt, a language model may generate continuations such as ``doctor'', ``nurse'', or ``teacher''. 
These responses reveal the model’s implicit associations between demographic attributes and occupations. 
For instance, a preference for ``nurse'' may reflect a gender stereotype associating women with caregiving roles. 
By composing prompts using these three components, HEIMAT systematically elicits and analyzes a model’s bias behavior across diverse contexts, forming the basis for subsequent debiasing.

Inspired by prior work~\citep{dong2024disclosure}, we construct heuristic prompts using a set of predefined templates to simulate realistic social scenarios while reducing manual prompt engineering cost. 
Each template generates a complete heuristic prompt by taking as input a neutral target word $W_t$, a feature word phrase $W_f$, and a bias type word $W_b$. 
For masked language models (e.g., BERT), an additional \texttt{[MASK]} token is included at the prediction position, whereas for autoregressive language models (e.g., LLaMA-2), the model is prompted to continue the given prefix.
We employ six templates in total to generate heuristic prompts in our experiments, ensuring that bias disclosure is grounded in realistic and socially meaningful settings while allowing the model to reveal its own inherent predispositions.

Given a heuristic prompt, we input the sequence into the model and extract the top-$k$ most probable tokens conditioned on the prompt context. 
For masked language models, this corresponds to the probability distribution over the masked position, while for decoder-only models, it corresponds to the next-token distribution given the prompt prefix. 
We denote the resulting token list as 
$WL = (w_1, w_2, \ldots, w_k)$, 
where each $w_i$ represents a demographic attribute implicitly associated with the preceding context. 
The set $WL$ thus captures the model’s most salient demographic associations under the heuristic prompt.

To further expand demographic coverage, we apply association prompts to each $w_i \in WL$ and prompt the model to generate $n$ related demographic terms, forming a substitution list $SL$:
\begin{equation}
\begin{aligned}
SL = [(w_1^1, w_2^1, \ldots, w_n^1), (w_1^2, w_2^2, \ldots, w_n^2), \\\ldots, (w_1^k, w_2^k, \ldots, w_n^k)].
\end{aligned}
\end{equation}

Using the substitution list, we construct a \emph{Context Prompt Set} (CPS) by minimally modifying the original heuristic prompt (as illustrated in Figure~\ref{fig:modify-prompts}). 
Specifically, we substitute the demographic attribute in the prompt with each term sampled from $SL$, while masking or abstracting the variable component of the feature words $W_f$. For autoregressive models, we reorder the sentence to place the next-token position at the end, allowing models to condition on the preceding demographic attributes.
This procedure ensures that all prompts within a CPS differ only in demographic attributes, while preserving identical contextual semantics.
Formally, we define the context prompt set as
\begin{equation}
CPS = (CP_1, CP_2, \ldots, CP_n),
\end{equation}
where each $CP_i$ denotes a context prompt. 
Each CPS is treated as a training unit in the subsequent debiasing fine-tuning stage.

\subsection{Debiasing Fine-tuning}

Given a language model \emph{M} and a context prompt \emph{CP}, the model induces a probability distribution over possible next tokens.
We define the prediction distribution \( p \) as:
\begin{equation}
    p(w) = M(w \mid CP),
\end{equation}
where \( w \) denotes a candidate token in the model’s output vocabulary.

Our goal is to ensure that the model produces consistent prediction distributions across context prompts that differ only in demographic attributes, while preserving the same semantic context.
Specifically, given a context prompt set \( CPS = (CP_1, CP_2, \ldots, CP_n) \), where each \( CP_i \) differs only in the substituted demographic term, we aim to minimize the discrepancy among their induced prediction distributions.

To achieve this, we adopt Jensen--Shannon Divergence~\citep{61115} as a symmetric and bounded measure of distributional difference, and fine-tune the model by minimizing the Jensen--Shannon Divergence loss (\(\mathcal{L}_{JSD}\)).
Given the set of prediction distributions \( PS = (p_1, p_2, \ldots, p_n) \), the loss is defined as:
\begin{equation}
\mathcal{L}_{JSD} = \frac{1}{n} \sum_{i=1}^{n} D_{KL}\left(p_i \;\middle\|\; \frac{1}{n} \sum_{j=1}^{n} p_j\right),
\end{equation}
where \( D_{KL} \) denotes the Kullback--Leibler Divergence.

By minimizing this loss during fine-tuning, the model is encouraged to align its output distributions across different contexts, thereby reducing biased associations while maintaining NLU capability.

\subsection{Language, Culture and Category Adaptation}

Although we provide only the heuristic and association prompts for English and French, the proposed method can be readily applied to other languages and cultural contexts. To adapt to a new culture, one simply prompts an LLM (e.g., DeepSeek) to regenerate the heuristic prompts following the same template structure, using culturally relevant feature words~(Examples can be found in \textbf{Appendix \ref{app:prompt-example}}.). This lightweight process replaces direct translation, ensuring that the biases being measured are those actually present in the target culture.

Furthermore, our templates are highly adaptable to various types of bias. By simply substituting the bias type word, such as changing 'gender' to 'race', we can effectively address and mitigate the model's biases related to race. This adaptability extends to a wide range of other bias categories, showcasing the versatility and broad applicability of our approach.

\section{Results and Analysis}
We describe the overall experimental results in this section and defer detailed configurations, including model choices, baseline implementations, datasets, and hyperparameter, to \textbf{Appendix \ref{app:experimental-setup}}.
\subsection{Quantitative Results}

\begin{table}[t]
  \centering
  \small
  \vspace{-0.2cm}
  \begin{tabular}{lcccc}
    \hline
    \hline
    \rowcolor{blue!15} & \multicolumn{2}{c|}{\textbf{CrowS-Pairs}} & \multicolumn{2}{c}{\textbf{StereoSet}} \\
    \rowcolor{blue!15} \multirow{-2}{*}{\textbf{Models}} & \textbf{Gender} & \textbf{Race} & \textbf{Gender} & \textbf{Race} \\
    \hline
    \rowcolor{blue!5}\textit{\textbf{BERT}} & 58.01 & 58.12 & 60.28 & 57.03 \\
    \ \ \ +CDA & 56.11 & \underline{56.70} & 59.61 & 56.73 \\
    \ \ \ +Dropout & 55.34 & 59.03 & 60.66 & 57.07 \\
    \ \ \ +INLP & 51.15 & 67.96 & 67.25 & 57.29 \\
    \ \ \ +Sent-Debias & 52.29 & 62.72 & 59.37 & 55.18 \\
    \ \ \ +Self-Debias & 52.29 & \underline{56.70} & 59.34 & 54.30 \\
    \ \ \ +Auto-Debias & 54.92 & 65.05 & 57.33 & 54.02 \\
    \ \ \ +FineDeb & 54.58 & 65.24 & \underline{53.27} & \textbf{50.82} \\
    \ \ \ +CCPA & 51.57 & - & 56.61 & - \\
    \ \ \ +ADELE & 54.20 & - & 46.27 & - \\
    \ \ \ +PromptDeb & \underline{50.31} & - & 59.80 & - \\
    \ \ \ +ChatGPT-Based & 51.68 & 57.20 & 71.80 & 59.60 \\
    \ \ \ +Bias Vector & 51.74 & 52.90 & - & - \\
    \rowcolor{gray!10} \ \ \ \textbf{+Ours} & \textbf{50.00} & \textbf{47.48} & \textbf{52.67} & \underline{53.74} \\
    \hline
    \hline
  \end{tabular}
  \caption{Comparison of debiasing performance of PLMs across different benchmarks. For CrowS-Pairs and StereoSet, scores closer to 50 are better. CCPA, PromptDeb and ADELE are specifically designed for gender debiasing; race-related scores are omitted. ``-'' indicates the method is not applicable to this dataset. (\textbf{Best}, \underline{Next Best}).}
  \label{tab:combined-res}
\end{table}

\begin{table}[t]
  \centering
  \small 
  \vspace{-0.2cm}
  \begin{tabular}{lcccc} 
    \hline
    \hline
    \rowcolor{blue!15} & \multicolumn{2}{c}{\textbf{CrowS-Pairs}} & \multicolumn{2}{c}{\textbf{StereoSet}} \\
    \rowcolor{blue!15} \multirow{-2}{*}{\textbf{Models}} & \textbf{Gender} & \textbf{Race} & \textbf{Gender} & \textbf{Race} \\
    \hline
    \rowcolor{blue!5}\textit{\textbf{Llama-2}} & 60.30 & 67.60 & 70.60& 63.00\\
    \ \ \ +CDA & 58.33 & 71.40 & 64.03& 67.24\\
    \ \ \ +INLP & 60.37 & 65.62 & 63.97& 62.50\\
    \ \ \ +Self-Debias& 54.87& 62.81&60.04& 63.49 \\
    \ \ \ +Synonym-KG& 59.20& 65.50& 67.50& 62.70\\
    \ \ \ +KGDebias& 56.90& 57.00& 61.20& 62.50\\
    \rowcolor{gray!10} \ \ \ \textbf{+Ours} & \textbf{54.07} & \textbf{55.64} & \textbf{55.45}& \textbf{53.17}\\
    \hline
    \rowcolor{blue!5}\textit{\textbf{GPT-2}} & 56.87& 59.69& 65.58& 61.63\\
    \ \ \ +CDA& 56.87& 60.66& 64.02& 57.31\\
    \ \ \ +INLP& 53.44& 59.69& 63.17& 60.00\\
    \ \ \ +Self-Debias& 56.11& \textbf{53.29}& 60.28& \underline{57.29}\\
    \ \ \ +Synonym-KG& 59.80& 55.80& 70.60& 59.80\\
    \ \ \ +KGDebias& 59.50& 55.40& 69.80& 58.30\\
    \ \ \ +PromptDeb& \underline{47.17}& -& \underline{55.10}& -\\
    \rowcolor{gray!10} \ \ \ \textbf{+Ours} & \textbf{52.52}& \underline{53.31}& \textbf{54.08}& \textbf{54.60}\\
    \hline
    \hline
  \end{tabular}
  \caption{Comparison of debiasing performance of LLMs across different benchmarks. For CrowS-Pairs and StereoSet, scores closer to 50 are better. (\textbf{Best}, \underline{Next Best}).}
  \label{tab:combined-res-llm}
\end{table}

\begin{table*}[t]
  \centering
  \small 
 
  \vspace{-0.2cm}
  \setlength{\tabcolsep}{3.8pt} 
  \begin{tabular}{llccccccccc}
    \hline
    \hline
    \rowcolor{blue!15} \textbf{Lang.} & \textbf{Models} & \textbf{Race} & \textbf{Gender} & \textbf{Socioeco.} & \textbf{Nationality} & \textbf{Religion} & \textbf{Age} & \textbf{Sexual} & \textbf{Apperance} & \textbf{Disability} \\
    \hline
    \rowcolor{blue!5} & \textit{\textbf{BERT}} & 58.12 & 58.01 & 59.93 & \textbf{62.94} & 71.39 & \textbf{55.23} & 67.92 & 63.46 & 61.68 \\
    & \ \ \ +Ours & \textbf{47.48} & \textbf{50.00} & \textbf{54.07} & 64.15 & \textbf{43.81} & 63.22 & \textbf{45.24} & \textbf{58.73} & \textbf{58.33} \\
    \cline{2-11}
    \rowcolor{blue!5} & \textit{\textbf{ALBERT}} & \textbf{51.36} & 57.25 & \textbf{60.47} & 51.57 & 59.05 & 66.52 & 75.00 & \textbf{46.03} & 86.67 \\
    & \ \ \ +Ours & 53.49 & \textbf{54.25} & 63.37 & \textbf{47.17} & \textbf{50.48} & \textbf{55.17} & \textbf{71.43} & 58.73 & \textbf{75.00} \\
    \cline{2-11}
    \rowcolor{blue!5} \multirow{-6}{*}{\textbf{EN}} & \textit{\textbf{TinyBERT}} & 68.22 & \textbf{50.76} & 68.02 & 54.72 & 31.43 & 58.62 & \textbf{47.62} & 47.62 & 56.67 \\
    & \ \ \ +Ours & \textbf{45.74} & 48.09 & \textbf{41.28} & \textbf{46.54} & \textbf{33.33} & \textbf{45.98} & 65.48 & \textbf{49.21} & \textbf{48.33} \\
    \hline
    \rowcolor{blue!5} & \textit{\textbf{CamemBERT}} & \textbf{53.04} & 57.01 & 59.69 & 62.45 & 65.22 & 53.33 & 52.75 & \textbf{58.33} & \textbf{62.12} \\
    & \ \ \ +Ours & 41.74 & \textbf{52.34} & \textbf{54.29} & \textbf{37.94} & \textbf{62.61} & \textbf{52.22} & \textbf{48.35} & 63.89 & 77.27 \\
    \cline{2-11}
    \rowcolor{blue!5} \multirow{-4}{*}{\textbf{FR}} & \textit{\textbf{FrALBERT}} & 56.74 & 47.66 & 58.16 & \textbf{60.47} & 72.17 & 38.89 & 81.32 & 40.28 & 42.42 \\
    & \ \ \ +Ours & \textbf{50.65} & \textbf{47.98} & \textbf{57.65} & 60.87 & \textbf{65.21} & \textbf{45.56} & \textbf{79.12} & \textbf{58.33} & \textbf{48.48} \\
    \hline
    \hline
  \end{tabular}
   \caption{Debiasing performance of HEIMAT on English (EN) and French (FR) CrowS-Pairs across bias categories. Notably, the models are only debiased in Gender and Race. Closer to 50 is better (\textbf{Best}). (Nat.: Nationality).}
  \label{tab:combined-crows}
\end{table*}

\begin{table}[t]
  \centering
  
  \vspace{-0.2cm}
  \begin{tabular}{lccc}
    \hline
    \hline
    \rowcolor{blue!15}\textbf{Models} & \textbf{Stereo} & \textbf{Anti-Stereo}& \textbf{Overall}\\
    \hline
    \rowcolor{blue!5}\textit{\textbf{BERT}}  & 61.09&56.88&60.48 \\
    \ \ \ +Ours     &\textbf{52.40}&\textbf{48.62}&\textbf{51.86}  \\
    \hline
    \rowcolor{blue!5}\textit{\textbf{ALBERT}}     &56.20&\textbf{60.09}&56.76  \\
    \ \ \ +Ours     &\textbf{55.35}&63.76&\textbf{55.56}   \\
    \hline
    \rowcolor{blue!5}\textit{\textbf{TinyBERT}}  & 58.39&\textbf{57.34}&58.16 \\
    \ \ \ +Ours     &\textbf{43.75}&61.29&\textbf{46.22}  \\
    \hline
    \hline
  \end{tabular}
  \caption{The performance of HEIMAT on three PLMs for the two most common categories in CrowS-Pairs. Closer to 50 is better~(\textbf{Best}).}
  \label{tab:3-models-cp-res}
\end{table}




To quantitatively evaluate the effectiveness of HEIMAT in mitigating social biases, we conduct experiments on the CrowS-Pairs, StereoSet, and SEAT benchmarks across gender and race dimensions using BERT.
As shown in Table~\ref{tab:combined-res}, HEIMAT consistently achieves the best or near-best performance, with scores closest to the ideal value of 50. In particular, on the CrowS-Pairs benchmark, HEIMAT reduces the Gender bias of BERT from 58.01 to 50.00 and achieves the lowest Race bias score (47.48), outperforming all competing baselines.
HEIMAT also achieves consistently strong performance on StereoSet, and yields the lowest effect sizes on the SEAT benchmark for both Gender and Race (reported in Table~\ref{tab:seat-res} of \textbf{Appendix \ref{app:additional-seat}}).

Similar trends are observed for LLaMA-2 and GPT-2, where HEIMAT achieves the lowest bias for both models, despite their reduced model capability. These results demonstrate that HEIMAT effectively mitigates demographic biases across different bias types, validating the robustness and generality of the proposed heuristic-style debiasing framework.


In contrast, several baselines show inconsistent behavior across datasets, improving bias under one metric while degrading under another. HEIMAT, however, achieves consistent improvements across all evaluated benchmarks, suggesting better robustness and generalization.




\subsection{Stereotype \& Anti-Stereotype Trade-off}


We further investigate the behavior of HEIMAT on whole CrowS-Pairs by jointly analyzing the overall stereotype tendency and its transfer effects across bias categories. As shown in Table~\ref{tab:3-models-cp-res}, HEIMAT consistently reduces Stereotype scores and moves the Overall scores of all three PLMs significantly closer to the ideal value of 50, indicating effective mitigation of dominant stereotypical bias across different model capacities. In particular, HEIMAT reduces the Overall score of BERT from 60.48 to 51.86 and that of TinyBERT from 58.16 to 46.22. We also observe increases in Anti-Stereotype scores for ALBERT and TinyBERT, suggesting a shift away from stereotypical preferences toward anti-stereotypical predictions, which is expected given the strong initial bias of these models and remains beneficial as reflected by the improved Overall scores.

Beyond the two debiased categories, Table~\ref{tab:combined-crows} further shows that debiasing conducted only on gender and race leads to consistent score shifts across multiple unseen bias categories, including socio-economic status, nationality, religion, and physical appearance. Such cross-category improvements indicate that different forms of social bias are intrinsically correlated in models probably, and that mitigating prominent biases can partially generalize to related dimensions. 

\subsection{Multilingual and Multicultural Debiasing}

\begin{figure}[t]
    \centering
    \scalebox{0.8}{
  \includegraphics[width=\columnwidth]{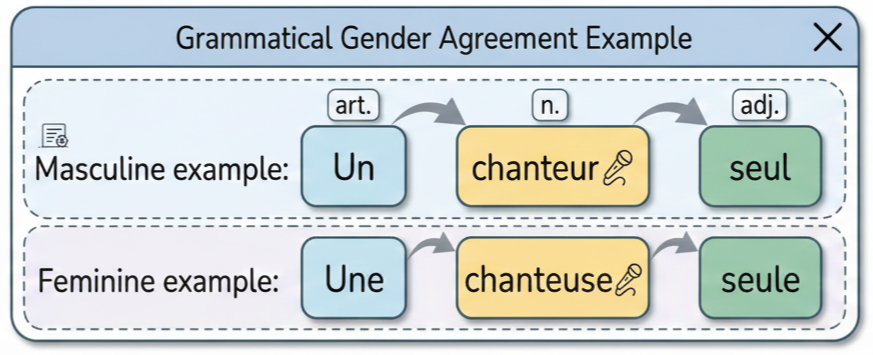}}
  \caption{Two examples of linguistic phenomenon of noun gender (meaning: “a lonely singer”).}
  \label{fig:gender}
\end{figure}

To evaluate the effectiveness of HEIMAT in multilingual and multicultural settings, we conduct debiasing experiments on two French PLMs, CamemBERT and FrALBERT, using the French version of CrowS-Pairs. The results are reported in Tables~\ref{tab:combined-crows}. When applying HEIMAT to French PLMs, the prompt templates are adapted to account for the grammatical structure of French, which differs from English in that nouns are associated with grammatical gender and require agreement with corresponding articles and adjectives. As illustrated in Fig.~\ref{fig:gender}, feminine and masculine forms exhibit systematic suffix variations, challenging fixed-template methods with static word lists~\citep{dong2024disclosure,zeng2024prompting} in cross-lingual and cross-cultural adaptation. Our method retains template structure but replaces static lists with LLM-generated content, offering both flexibility and clarity.

Despite this additional linguistic complexity and cultural difference, HEIMAT consistently reduces bias on both models, bringing the Overall scores closer to the ideal value of 50. Specifically, HEIMAT reduces CamemBERT’s Overall score from 57.36 to 50.69 and improves FrALBERT’s Anti-Stereotype balance by aligning its predictions toward a more neutral distribution. Therefore, these results demonstrate that HEIMAT effectively mitigates bias in French PLMs and can be readily extended to other cultures and languages

\subsection{Qualitative Bias Analysis}

\begin{figure}[t]
    \centering
    \scalebox{0.95}{
  \includegraphics[width=\columnwidth]{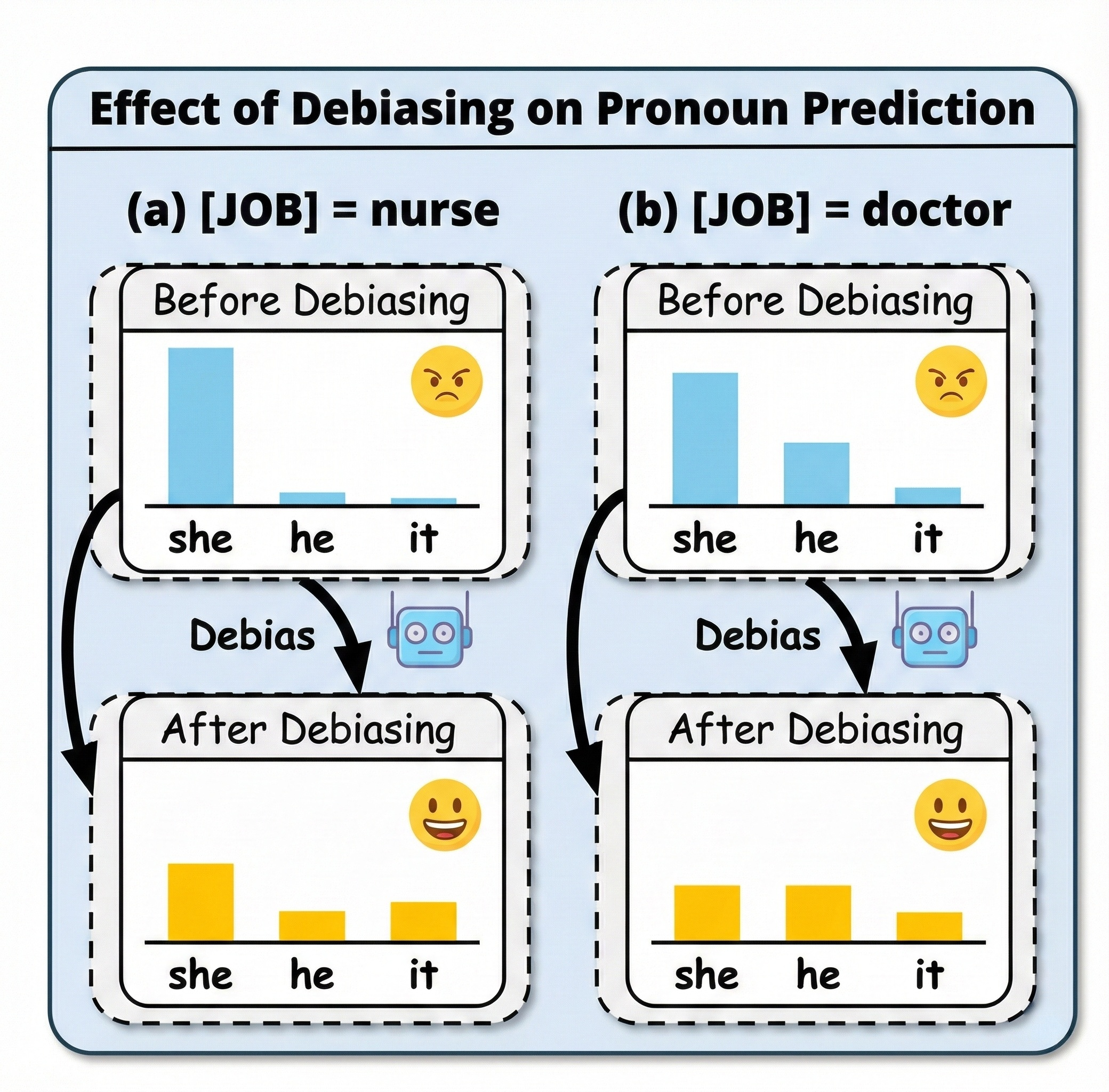}}
  \caption{Visualization of the predicted probabilities of gender pronoun for “nurse” and “doctor” on ALBERT before and after debiasing with HEIMAT.}
  \label{fig:case-study}
\end{figure}

Figure~\ref{fig:case-study} presents a qualitative comparison of the prediction probability distributions produced by the original and debiased ALBERT under a controlled prompt setting. Specifically, we use the prompt “\emph{This [JOB] works in a famous hospital and [MASK] is professional}” and substitute “nurse” and “doctor” for [JOB] to examine gender pronoun predictions. In the original ALBERT, a pronounced disparity is observed: when the job is “nurse”, the model assigns substantially higher probability to the feminine pronoun “she”, whereas for “doctor”, the masculine pronoun “he” dominates, revealing strong occupation–gender associations.

In contrast, the debiased model exhibits markedly more balanced prediction distributions. The probability gap between “he” and “she” is significantly reduced for both occupations, indicating that the model’s predictions become less sensitive to gender stereotypes associated with specific jobs. This qualitative analysis demonstrates that HEIMAT effectively attenuates demographic-conditioned prediction shifts at the distributional level, aligning with the quantitative improvements observed on bias benchmarks.

Similar phenomenon can be observed in Llama-2, as shown in Table \ref{box:llama2-example} Before debiasing, the model tends to complete prompts involving racial identifiers with narratives of violence and victimization. After debiasing, while negative outcomes may still appear, the generated continuations no longer systematically attribute harm to external agents conditioned on race, suggesting a reduction in identity-conditioned violent associations.

\begin{table}[t]
\begin{tcolorbox}[
  colback=gray!5,
  colframe=black,
  boxrule=0.5pt,
  arc=2mm,
  title={Model Generations Before and After Debiasing},
  fonttitle=\bfseries\small
]
\small
\textbf{Prompt:}  
On Sunday, this Black man like...
\vspace{0.5em}

\textbf{LLaMA-2 (Before Debiasing):}  
On Sunday, this black man like myself \textcolor{red}{was killed by a white police officer} in the city of Philadelphia. He was killed for being ...
\vspace{0.5em}

\textbf{LLaMA-2 (After Debiasing):}  
On Sunday, this black man like has \textcolor{cyan}{a serious accident and he is in a coma}. He is on life support. He is brain ...

\end{tcolorbox}
  \caption{Example generations of LLaMA-2 before and after debiasing.}
\label{box:llama2-example}
\end{table}

\subsection{Analysis of NLU Tasks Performance}
Since HEIMAT mitigates bias via fine-tuning, it is important to examine its potential impact on the model’s NLU performance~\citep{meade2022empirical}. We therefore evaluate HEIMAT on standard NLU benchmarks to assess whether bias reduction is achieved while preserving NLU capability.

\begin{table*}
  \centering
  \small
  \scalebox{0.95}{
  \begin{tabular}{lcccccccccc}
    \hline
    \hline
    \rowcolor{blue!15}\textbf{Models} & \textbf{CoLA} & \textbf{SST-2}& \textbf{MRPC}&\textbf{QNLI}&\textbf{QQP}&\textbf{RTE}&\textbf{MNLI}&\textbf{STS-B}&\textbf{WNLI}&\textbf{Avg.}\\
    \hline

    \rowcolor{blue!5}\textit{\textbf{BERT}}& 56.78 & \textbf{93.35} & \textbf{89.54} & \underline{91.51} & \underline{88.06} & 64.62 & 84.76 & 88.24 & \textbf{56.34} & \textbf{79.24} \\ 
    \ \ \ +CDA & 55.97 & 92.32 & 87.22 & 90.84 & 87.85 & 63.29 & 84.84 & 88.43 & 53.66 & 78.27 \\ 
    \ \ \ +Dropout & 50.87 & 92.09 & 88.22 & 91.49 & 88.02 & 62.29 & 84.78 & 87.87 & 51.66 & 77.48 \\ 
    \ \ \ +INLP & 56.50 & 92.66 & \underline{89.23} & 91.38 & 87.94 & \underline{65.34} & 84.78 & \textbf{88.73} & 54.93 & 79.05 \\
    \ \ \ +Sent-Debias & 55.72 & \underline{93.12} & 88.81 & 91.54 & 87.88 & 63.90 & \textbf{84.94} & 88.23 & \textbf{56.34} & 78.94 \\ 
    \ \ \ +Auto-Debias & \underline{57.01} & 92.89 & 88.54 & \textbf{91.65} & 87.92 & 64.62 & \underline{84.91} & 88.43 & 40.85 & 77.42 \\ 
    \ \ \ +CCPA & 55.91 & 93.09 & 88.65 & 91.42 & 87.98 & 64.93 & 84.73 & \underline{88.44} & \underline{55.66} & 78.98 \\
    \ \ \ +Ours&\textbf{57.40}&91.80&88.84&91.09&\textbf{88.54}&\textbf{65.51}&84.72&87.66&\textbf{56.34}&\underline{79.09}\\
    \hline

    \rowcolor{blue!5}\textit{\textbf{ALBERT}}&55.61&\underline{92.32}&90.85&91.21&\underline{88.99}&\underline{72.56}&\underline{85.38}&89.39&39.44 &78.42\\
    \ \ \ +Dropout & 46.83 & 91.86 & 89.83 & 90.99 & 86.90 & 55.71 & \underline{85.13} & 88.68 & 52.34 & 76.47 \\ 
    \ \ \ +Sent-Debias & \underline{56.37} & 92.31 & \textbf{91.65} & \underline{91.92} & 87.56 & 67.87 & 84.96 & \underline{90.46} & 33.80 & 77.43 \\ 
    \ \ \ +Auto-Debias & \textbf{56.69} & \textbf{93.35} & \underline{91.61} & \textbf{91.94} & 87.41 & \textbf{73.29} & 79.25 & \underline{90.48} & \textbf{54.93} & \textbf{79.88} \\ 
    \ \ \ +Ours     &55.42&92.15&90.87&91.02&\textbf{89.74}&69.26&85.01&90.35&\underline{53.28}&\underline{79.67}\\
    \hline

    \rowcolor{blue!5}\textit{\textbf{TinyBERT}}    &\textbf{51.34}&\textbf{92.12}&\underline{86.69}&\textbf{90.15}&\textbf{70.64}&\underline{70.16}&\textbf{84.74}&\underline{83.42}&\underline{56.28}&\textbf{76.28}\\
    \ \ \ +Ours~~          &\underline{48.48}&\underline{91.62}&\textbf{87.43}&\underline{90.08}&\underline{70.03}&\textbf{73.31}&\underline{84.33}&\textbf{84.70}&\textbf{56.30}&\underline{76.25}  \\
    \hline

    \rowcolor{blue!5}\textit{\textbf{LLaMA-2}} &\textbf{18.64}& \textbf{75.73} &\underline{66.37}& \underline{58.57}& \underline{31.39} &\textbf{59.32}& \textbf{88.76} &\textbf{27.51}& \underline{44.50} &\textbf{52.31} \\
    \ \ \ +Ours~~          
    
    &\underline{18.34}&\underline{75.40}&\textbf{67.43}&\textbf{59.01}&\textbf{31.88}&\underline{58.40}&\underline{86.24}&\underline{27.10}&\textbf{45.16}&\underline{52.00}  \\
    
    \rowcolor{blue!5}\textit{\textbf{GPT-2}} & \textbf{76.84} & \textbf{91.20} & \underline{80.41} & \textbf{88.37} & \textbf{89.66} & \textbf{65.30} & \underline{82.18} & \textbf{37.10} & \textbf{47.79} & \textbf{73.21} \\
    \ \ \ +Ours~~          &\underline{75.41}&\underline{90.99}&\textbf{81.00}&\underline{87.95}&\underline{89.02}&\underline{64.73}&\textbf{82.20}&\underline{37.08}&\underline{45.72}&\underline{72.68}  \\
    
    \hline
    \hline
  \end{tabular}}
    
  \caption{Performance comparison of debiasing on the GLUE Benchmark. We chose \textbf{F1 Score} as the metric on the MRPC and QQP and \textbf{Spearman Correlation} as the metric on STS-B. LLaMA-2 is evaluated in a zero-shot setting, while the other models are fine-tuned on downstream task(\textbf{Best}, \underline{Next Best}).}
  \label{tab:nlu-en}
\end{table*}

\noindent\textbf{NLU Tasks Performance of English PLMs.} To evaluate whether HEIMAT preserves the NLU capability of PLMs, we conduct experiments on the GLUE benchmark, with the results reported in Table~\ref{tab:nlu-en}. The results indicate that HEIMAT maintains NLU performance at a level comparable to the original models while remaining competitive with existing debiasing methods. Specifically, for BERT, HEIMAT achieves an average GLUE score of 79.09, which is close to the original score of 79.24, with performance on tasks such as QQP and RTE even slightly improved. For ALBERT, HEIMAT attains an average score of 79.67, comparable to the base model (78.42) and competitive with the strongest baseline. Similarly, for TinyBERT, LLama-2 and GPT-2, the average GLUE score remains essentially unchanged, indicating minimal performance degradation despite the reduced model capacity. Therefore, these results demonstrate that HEIMAT effectively mitigates bias while preserving the core NLU capabilities of PLMs.

\begin{table}
  \centering

  \resizebox{\linewidth}{!}{
  \begin{tabular}{lcccc}
  \hline
    \hline
    \rowcolor{blue!15}\textbf{Models} & \textbf{PAWS-X} & \textbf{XNLI}& \textbf{~CLS~}&\textbf{~Avg.~}\\
    \hline
    \rowcolor{blue!5}\textit{\textbf{CamemBERT}} & 91.24&81.16&94.75&89.05 \\
    \ \ \ +Ours     & 91.50&81.47&94.59&\textbf{89.18} \\
    \hline
    \rowcolor{blue!5}\textit{\textbf{FrALBERT}}     &83.44&71.06&80.42&\textbf{78.30}  \\
    \ \ \ +Ours     &83.87&70.98&79.74&78.19   \\
    \hline
    \hline
  \end{tabular}}
    \caption{The performance of our method on FLUE Benchmark.}
  \label{tab:2-models-flue}
\end{table}

\noindent\textbf{NLU Tasks Performance of French PLMs.} To evaluate whether HEIMAT preserves the NLU capability of French PLMs, we conduct experiments on the FLUE benchmark, with results reported in Table~\ref{tab:2-models-flue}. The results show that HEIMAT maintains comparable NLU performance on French PLMs after debiasing. Specifically, for CamemBERT, HEIMAT achieves an average score of 89.18, which is slightly higher than the baseline score of 89.05, with consistent performance on PAWS-X and XNLI. For FrALBERT, although the scores on XNLI and CLS decrease marginally (from 71.06 to 70.98 and from 80.42 to 79.74, respectively), the overall average remains largely unchanged (78.30 vs. 78.19). Therefore, these results demonstrate that HEIMAT effectively mitigates bias in French PLMs without causing significant degradation in their NLU performance.


\section{Conclusion} 
\label{sec:bibtex}
In this work, we propose HEIMAT, a heuristic-style automatic debiasing framework for LMs that leverages a small set of heuristic templates and uses DeepSeek to generate heuristic prompts to disclose social biases, then mitigates model biases through JSD-based fine-tuning. By avoiding reliance on external corpora, HEIMAT can be readily extended to different bias categories, languages, and cultures, which we validate through experiments on both PLMs and LLMs in different languages. Extensive results across multiple benchmarks demonstrate that HEIMAT achieves effective bias mitigation with minimal impact on general performance, while also revealing intrinsic connections among different bias types that enable broader generalization.

\section*{Limitations}
We acknowledge that not all stereotypical associations are inherently harmful, as some reflect widely shared cultural knowledge. For example, Christians are more likely to appear in churches, while people from other religions have a lower probability. Distinguishing between harmful social biases and benign common-sense patterns remains an open challenge, and we leave this important issue for future work.


\section*{Ethics Statement}

The bias evaluation datasets we used are available publicly, and we are not responsible for any content that contains stereotypes and biases in these datasets. Our work focuses on reducing model bias and using them to evaluate the performance of our method.

\bibliography{custom}

\appendix


\section{Prompt Examples}
\label{app:prompt-example}
In Table \ref{tab:appendix-examples}, we present the heuristic prompt samples in English and French respectively.

The structures of the six English heuristic templates we use are reported in Table \ref{tab:appendix-templates}. Among them, \( W_{t1} \) and \( W_{t2} \) are two different target word. The [VERBS] represents cognitive verbs describing thinking, perception, understanding, or other psychological processes, such as 'think', 'know' and 'guess'. The words mentioned above, except \( W_f \) about professions, are generated by ChatGPT.

We have listed the association prompts for the model to generate new English target words in similar domains in Table \ref{tab:appendix-wl-prompts}. We did not find words by calculating cosine similarity, selecting the few words with the smallest Euclidean distance from the vocabulary, or using clustering because these methods tend to result in poor diversity of the words obtained, and they often include words in irrelevant domains with too significant semantic differences.

\section{Experimental Setup}
\label{app:experimental-setup}
\subsection{Datasets}
\noindent We evaluate the effectiveness of HEIMAT across both PLMs and large language models (LLMs). For English PLMs, we utilize several representative datasets, including CrowS-Pairs, StereoSet, SEAT, and the GLUE benchmark. To assess the method's cross-lingual generalizability, we further extend our evaluation to French models using French CrowS-Pairs and the FLUE benchmark. For LLMs, following~\citet{touvron2023llama}, we evaluate English debiasing performance on CrowS-Pairs (via perplexity) and StereoSet. In all experiments, GLUE and FLUE serve as the primary metrics for assessing Natural Language Understanding (NLU) performance post-debiasing.

\subsection{Language Models}

\noindent\textbf{Pre-trained Language Models}
Following previous works, we evaluate our method with two famous English PLMs~(bert-base-uncased~\citep{devlin-etal-2019-bert}; albert-base-v2~\citep{lan2020albert};TinyBERT-General-6L-768D~\citep{jiao2020tinybert}; and two famous French PLMs~(camembert-base~\citep{martin-etal-2020-camembert}; fralbert-base~\citep{cattan2022usability}).

\noindent\textbf{Large Language Models}
For LLMs, we apply our debiasing method to Llama2-7b-hf~\citep{touvron2023llama} and GPT-2 medium~\citep{radford2019language}.

\subsection{Baselines}
We choose Counterfactual Data Augmentation (CDA)~\citep{lu2019gender}, Sent-Debias~\citep{liang2020debiasing}, Iterative Nullspace Projection (INLP)~\citep{ravfogel2020null}, Dropout~\citep{webster2021measuring}, Self-Debias~\citep{schick2021selfdiagnosis}, Auto-Debias~\citep{guo-etal-2022-auto}, CCPA~\citep{li2023prompt}, FineDeb~\citep{saravanan2023finedeb}, ChatGPT-Based~\citep{han2024chatgpt}, ADELE~\citep{zakizadeh-pilehvar-2025-blind}, Synonym-KG, KGDebias~\citep{ma-etal-2024-debiasing}, PromptDeb~\citep{zeng2024prompting} and Bias Vector~\citep{shirafuji-etal-2025-bias} as our baselines.

\subsection{Hyperparameters}
All of the experiments were conducted on an A100 with 80GB of memory, utilizing the Adam optimizer~\citep{loshchilov2019decoupled} with an initial learning rate set to 5e-6, a batch size of 8. We generate 20000 heuristic prompts in the first step and collected the top 5 words with the highest probabilities. The other baselines are ran with their default settings. 


\section{Complete Evaluation Results for SEAT}
\label{app:additional-seat}
Table \ref{tab:seat-res} records the effect sizes for tests in SEAT. The result shows that HEIMAT effectively reduces social bias in PLMs.

\begin{table}[t]
  \centering
  \vspace{-0.2cm}
  \begin{tabular}{lcc}
    \hline
    \hline
    \rowcolor{blue!15}\textbf{Models} & \textbf{Gender} & \textbf{Race} \\
    \hline
    \rowcolor{blue!5}\textit{\textbf{BERT}} & 0.621 & 0.716 \\
    \ \ \ +CDA & 0.722 & 0.685 \\
    \ \ \ +Dropout & 0.765 & \underline{0.580} \\
    \ \ \ +INLP & \underline{0.204} & 0.675 \\
    \ \ \ +Sent-Debias & 0.430 & 0.610 \\
    \ \ \ +Self-Debias & 0.621 & 0.716 \\
    \ \ \ +Auto-Debias & 0.671 & 0.663 \\
    \ \ \ +FineDeb & 0.360 & 0.620 \\
    \ \ \ +CCPA & 0.249 & - \\
    \ \ \ +Bias Vector & 0.653 & 0.646 \\
    \rowcolor{gray!10} \ \ \ \textbf{+Ours} & \textbf{0.183} & \textbf{0.237} \\
    \hline
    \hline
  \end{tabular}
  \caption{Debiasing performance on the SEAT benchmark. Absolute effect sizes closer to 0 indicate lower bias. Methods marked with ``-'' are not applicable to the corresponding setting. (\textbf{Best}, \underline{Next Best}).}
  \label{tab:seat-res}
\end{table}

\section{How Different Setups Impact the Result}
\label{app:diffsetup}
We conduct an experiment on BERT after debiasing in gender and race domain to explore the impact of different number of heuristic prompts and word usage proportion on the debiasing performance. The results can be found in Table \ref{tab:ablation-score}. Overall, as the number of heuristic prompts and the UR increase, our method also performs better in the Overall Score of CrowS-Pairs.

\section{Time and Additional Content Required by Several Methods}
\label{app:ablation}
Table \ref{tab:time} records the time and additional content required by HEIMAT~(Ours), Auto-Debias, CDA, Self-Debias and FineDeb to complete a debiasing task in our device, among which the records regarding FineDeb are derived from the original paper.
\label{app:time}

\begin{table*}[t]
  \centering
  \scalebox{0.85}{
  \begin{tabular}{p{0.12\textwidth}p{0.12\textwidth}p{0.72\textwidth}}
  \hline
  \textbf{Language} & \textbf{Category} & \textbf{Sentence} \\
  \hline

  \multirow{2}{*}{English}
  & Gender & A person who has been working as a long-haul truck driver for twenty years, and this person's gender is widely described as [MASK]. \\
  & Race & A person who is the CEO of a major tech company, yet this person's race is still sometimes described as [MASK].\\
  \hline

  \multirow{2}{*}{French}
  & Gender & Une personne qui travaille comme conducteur ou conductrice de poids lourd sur les routes depuis vingt ans, et le genre de cette personne est généralement décrit comme [MASK]. \\
  & Race & Une personne qui est PDG d'une grande entreprise technologique, mais l'origine de cette personne est encore parfois décrite comme [MASK]. \\
  \hline

  \end{tabular}}
  \caption{Heuristic prompt examples in English and French across different social attributes.}
  \label{tab:appendix-examples}
\end{table*}

\begin{table*}
  \centering
  \begin{tabular}{lll}
  \hline
    \textbf{No.} & \textbf{Template}&\\
    \hline

    1&~~[\( W_{t1} \)] [\( W_f \) about hobbies] &~~~and [\( W_{t2} \)] [VERBS] [\( W_{t1} \)]'s [\( W_b \)] is [MASK].\\
    2&~~[\( W_{t1} \)] [\( W_f \) about adjectives] &~~~and [\( W_{t2} \)] [VERBS] [\( W_{t1} \)]'s [\( W_b \)] is [MASK].\\
    3&~~[\( W_{t1} \)] [\( W_f \) about professions] &~~~and [\( W_{t2} \)] [VERBS] [\( W_{t1} \)]'s [\( W_b \)] is [MASK].\\
    4&~~[\( W_{t1} \)] [\( W_f \) about hobbies] &~~~and [\( W_{t1} \)]'s [\( W_b \)] is [VERBS] as [MASK].\\
    5&~~[\( W_{t1} \)] [\( W_f \) about adjectives] &~~~and [\( W_{t1} \)]'s [\( W_b \)] is [VERBS] as [MASK].\\
    6&~~[\( W_{t1} \)] [\( W_f \) about professions] &~~~and [\( W_{t1} \)]'s [\( W_b \)] is [VERBS] as [MASK].\\
    \hline
  \end{tabular}
  \caption{The six heuristic prompt templates used in our method. }
  \label{tab:appendix-templates}
\end{table*}

\begin{table*}
  \centering
  \scalebox{0.93}{
  \begin{tabular}{lll}
  \hline
    \textbf{No.} & \textbf{Template}&\\
    \hline
    1&The word [MASK] has an opposite meaning to \{w\} and is often used as an opposite word.\\
    2&The synonym of word \{w\} is [MASK], which has a similar meaning to \{w\}.\\
    3&The word that has a similar meaning to {w} is [MASK].\\
    4&The opposite word to \{w\} is [MASK].\\
    5&The meaning of the word \{w\} and the word [MASK] is opposite.\\
    6&In this context, word \{w\} and word [MASK] have completely opposite meanings.\\
    7&There are many connections between the word \{w\} and the word [MASK].\\
    \hline
  \end{tabular}}
  \caption{Association prompt templates used in the same bias categories. "{w}" is the placeholder.}
  \label{tab:appendix-wl-prompts}
\end{table*}

\begin{table*}
  \centering
  \scalebox{0.99}{
    \begin{tabular}{cccc}
      \hline
      \multirow{2}{*}{\textbf{The Number of Heuristic Prompts}} & \multicolumn{3}{c}{\textbf{Overall Score}} \\
      \cline{2-4}
      & \textbf{UP = 1/4} & \textbf{UP = 1/2} & \textbf{UP = 1} \\
      \hline
      5000 & 61.21 & 60.47 & 57.63 \\
      10000 & 59.81 & 59.15 & 57.03 \\
      15000 & 59.92 & 57.63 & 54.25 \\
      20000 & 55.35 & 54.97 & 51.86 \\
      \hline
    \end{tabular}}
  \caption{Overall Score of CrowS-Pairs on BERT processed with HEIMAT with the different number of heuristic prompts and word usage proportions (UP) of given words mentioned in Section \ref{bias-disclosure}.}
  \label{tab:ablation-score}
\end{table*}


\begin{table*}
  \centering
  \scalebox{0.97}{
  \begin{tabular}{lp{0.3\textwidth}p{0.55\textwidth}}
    \hline
    \textbf{Methods} & \textbf{Time Required (hours)} & \textbf{Additional Content Required} \\
    \hline
    Auto-Debias & 20 & Word lists of demographic words and 5000 common words. \\
    FineDeb & 41 & Word lists of demographic words. \\
    CDA & Very long (requiring retraining with \( D_c \)) & New dataset \( D_c \) with counterfactual examples. \\
    Self-Debias & - & Original and biased prompts using the RealToxicityPrompts dataset and Perspective API. \\
    Ours & 0.5 & Heuristic, association and context prompts. \\
    \hline
  \end{tabular}}
  \caption{Time spent and additional content required for several debiasing methods}
  \label{tab:time}
\end{table*}

\end{document}